\documentclass[runningheads]{llncs}

\usepackage{accvabbrv}
\usepackage{accv}

\usepackage{graphicx}
\usepackage{booktabs}

\usepackage[accsupp]{axessibility}  % Improves PDF readability for those with disabilities.

\usepackage[pagebackref,breaklinks,colorlinks,citecolor=accvblue]{hyperref}
\usepackage{orcidlink}

\begin{document}

\title{KAN-Based Fusion of Dual-Domain for Audio-Driven Facial Landmarks Generation} 

\titlerunning{KFusion of Dual-Domain}

\author{Hoang-Son Vo-Thanh\inst{1}\orcidlink{0009-0001-3278-727X} \and
Quang-Vinh Nguyen\inst{1} \and
Soo-Hyung Kim\inst{1}$^*$\orcidlink{0000-0002-3190-0293}}

\authorrunning{Son et al.}

\institute{Department of Artificial Intelligence Convergence, Chonnam National University, Korea  \\
\email{\{hoangsonvothanh,vinhbn28,shkim\}@jnu.ac.kr}}

\maketitle

% --------------------ABTRACT-------------------------
\begin{abstract}
    % Overemphasizing image information processing instead of striving to build accurate landmarks from audio is not an optimal direction that current research on the Audio-driven emotional talking face problem. The synthesis of emotional information from speech landmarks should be thoroughly handled rather than just focusing on mouth shapes. To address these limitations, we introduce the ABC model in this paper. This model focuses on constructing the entire emotional face landmarks based on the voice. The model synthesizes information in two independent domains: the content domain, which focuses on processing emotional information, and the global domain, which handles information about mouth shapes and facial expressions. The results demonstrate the effectiveness of preserving facial expressions. Our method has over-performed when quantitative compared to related methods, especially on the mouth area landmarks.
    Audio-driven talking face generation is a widely researched topic due to its high applicability. Reconstructing a talking face using audio significantly contributes to fields such as education, healthcare, online conversations, virtual assistants, and virtual reality. Early studies often focused solely on changing the mouth movements, which resulted in outcomes with limited practical applications. Recently, researchers have proposed a new approach of constructing the entire face, including face pose, neck, and shoulders. To achieve this, they need to generate through landmarks. However, creating stable landmarks that align well with the audio is a challenge. In this paper, we propose the KFusion of Dual-Domain model, a robust model that generates landmarks from audio. We separate the audio into two distinct domains to learn emotional information and facial context, then use a fusion mechanism based on the KAN model. Our model demonstrates high efficiency compared to recent models. This will lay the groundwork for the development of the audio-driven talking face generation problem in the future.

    % Audio-driven talking face generation là một đề tài đang được nghiên cứu rộng rãi vì độ ứng dụng cao của nó. Việc tái tạo khuôn mặt đang nói chuyện bằng audio sẽ góp phần rất lớn vào các lĩnh vực như education, healthcare, online conversations, virtual assistants, and virtual reality. Những nghiên cứu ban đầu thường chỉ nhắm đến việc tập trung thay đổi chuyển động của khuôn miệng, điều này đem đến những kết quả không có tính ứng dụng cao. Gần đây, các nhà nghiên cứu đã đề xuất hướng giải quyết mới là xây dựng toàn bộ khuông mặt, bao gồm face pose, neck và vai. Để làm được điều đó họ cần phải generate thông qua các landmarks. Tuy nhiên để tạo được các landmarks ổn định, phù hợp với âm thanh là một thử thách. Trong bài viết này chúng tôi sẽ đề xuất mô hình KFusion of Dual-Domain, một mô hình mạnh mẽ giúp xây dựng các landmarks từ audio. Chúng tôi phân tách âm thanh thành 2 miền riêng biệt đề học tập thông tin cảm xúc, và bối cảnh khuôn mặt, sau đó sử dụng cơ chế hợp nhất dựa trên mô hình KAN. Mô hình của chúng tôi cho thấy độ hiệu quả rất cao khi so sánh với các mô hình gần đây. Đây sẽ là tiền đề cho sự phát triển của bài toán audio-driven talking face generation trong tương lai.
\end{abstract}
 
% --------------------ABTRACT-------------------------

% --------------------CONTENT-------------------------
\definecolor{accvblue}{rgb}{0,0,1.0}

\section{Introduction}
\label{sec:intro}

\begin{figure}[tb]
  \centering
  \includegraphics[width=\textwidth]{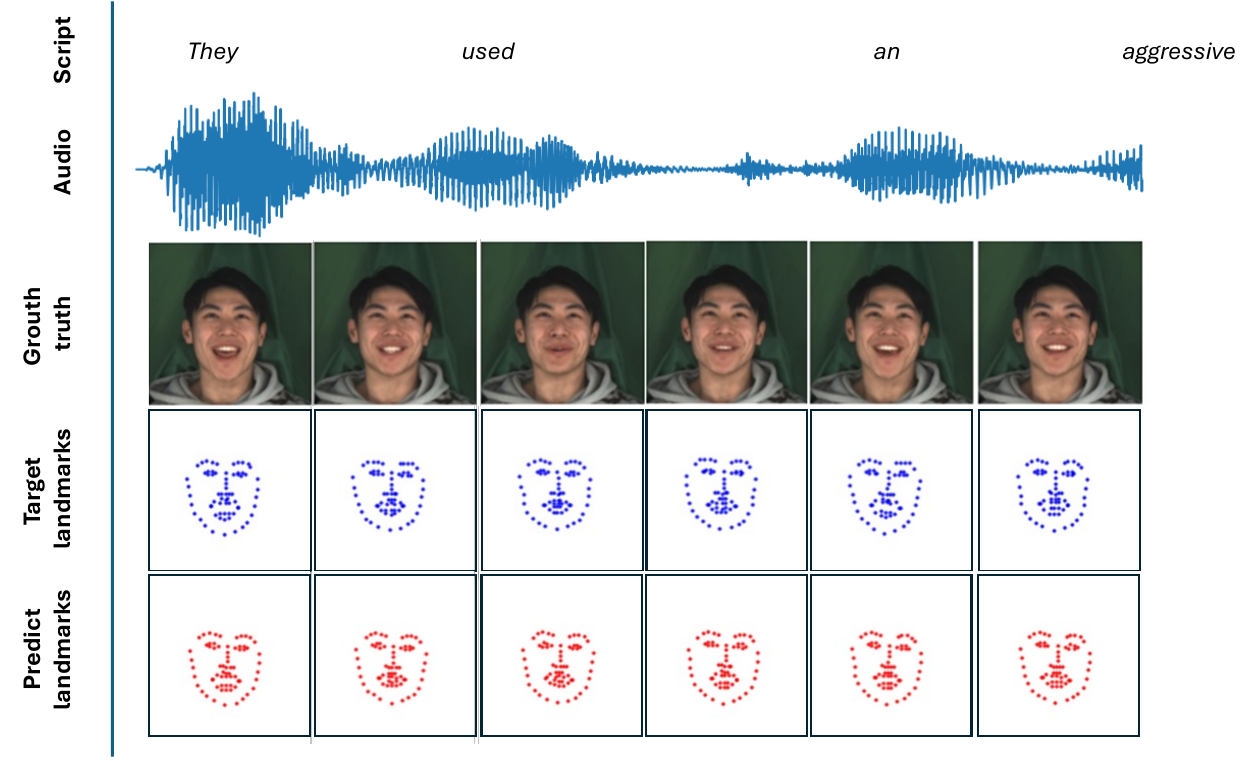}
  \caption{Overview of the problem we will address in this paper. We extract landmarks (in blue) from the ground truth video. Our model has learned to generate a sequence of landmarks (in red) from the audio.
  }
  % Tổng quan bài toán chúng tôi sẽ giải quyết trong bài báo này. Trong đó, chúng tôi extract landmarks (màu xanh dương) từ Grouth truth video. Mô hình của chúng tôi đã học cách kiến trúc lên sequence of landmarks (màu đỏ) từ audio.
  \label{fig:example}
\end{figure}

Audio-driven talking face generation is a rapidly evolving research field, gaining significant attention in recent years. This involves generating realistic human facial images and expressions corresponding to a given audio input. It also presents a powerful technology with promising applications in various domains, including education, healthcare, online conversations, virtual assistants, and virtual reality \cite{pataranutaporn2021ai}. It has the potential to significantly enhance user experiences by enabling more natural and engaging communication across diverse scenarios, surpassing the limitations of text-based or audio-only interactions.

To generate facial movements and expressions driven by audio, we first need a photo to determine identity. Current published research generally develops along two branches. One focuses on generating the movements of components such as the mouth and eyes and replacing them with the identity photo \cite{chung2017you,zhou2019talking, kr2019towards, prajwal2020lip, cheng2022videoretalking, zhou2020makelttalk} and they were yielded notable results. However, merely changing the movements of the mouth or eyes while retaining elements such as face pose or orientation makes the results unnatural. Therefore, the other branch reconstructs the face by building a sequence of facial landmarks and then generating the image based on these landmarks \cite{suwajanakorn2017synthesizing, ji2021audio, chen2019hierarchical, sinha2022emotion}. This branch is currently receiving more attention from researchers.

% Để tái tạo lại chuyển động và biểu cảm của khuôn mặt bằng âm thanh trước hết chúng ta cần có một tấm ảnh để xác định danh tính. Các nghiên cứu đã được công bố hiện nay thường sẽ phát triển theo hai nhánh, một sẽ tập trung generate ra chuyển động của các bộ phận như miệng, mắt và thay thế vào hind ảnh định danh []. Hướng tiếp cận này đã cho ra những kết quả đáng ghi nhận. Tuy nhiên, chỉ thay đổi chuyển động của miệng hay mắt mà giữ lại các yếu tố như face pose hay hướng của mặt làm cho kết quả không được tự nhiên. Bên cạnh đó, nhánh còn lại tập sẽ kiến trúc lại khuôn mặt bằng cách xây dựng chuỗi landmark của khuôn mặt sau đó generate ra hình ảnh dựa trên các landmarks []. Hiện nay hướng tiếp cận này đang được quan tâm nhiều hơn.

Those methods have demonstrated the effectiveness of this approach. The results have shown smooth movements. However, there are still some limitations, such as the mouth movements not matching the speech. Constructing a sequence of landmarks with movements that align with the dialogue is a challenge for current researchers.

% Các phương pháp ấy đã cho thấy độ hiệu quả của hướng tiếp cận này. Kết quả đã có thể chuyển động mượt mà tuy nhiên vẫn còn một số hạn chế như chuyển động của miệng không khớp với thoại. Để xây dựng được chuỗi các landmark với chuyển động khớp với hội thoại là một thử thách đối với các nhà nghiên cứu hiện nay.

In this paper, we propose a solution that focuses on constructing landmark sequences from speech. To effectively extract information from audio, we suggest processing the information in two domains and using a combination of LSTM \cite{sak2014long} and Transformer \cite{vaswani2017attention} models to extract features. Additionally, to combine information from the two data domains, we propose a fusion method based on the KAN model \cite{liu2024kan}. Therefore, the title given is ``KAN-Based Fusion of Dual-Domain for Audio-Driven Facial Landmarks''.

% Trong bài viết này, chúng tôi sẽ đề xuất một giải pháp tập trung vào việc xây dựng chuỗi landmark từ giọng nói. Để có thể trích xuất thông tin từ âm thanh một cách hiệu quả nhất, chúng tôi đề xuất xử lý thông tin theo hai miền và sử dụng kết hợp mô hình LSTM và Transformer để trích xuất đặc trưng. Bên cạnh đó, để kết hợp thông tin từ 2 miền dữ liệu, chúng tôi đề xuất phương pháp hợp nhất dựa trên mô hình KAN []. Vì vậy bài viết này sẽ có tên KAN-Based Fusion of Dual-Domain for Speech-Driven Facial Landmarks.

Our contribution will be summarized as follows:
\begin{itemize}
  \item Proposing the KFusion of Dual-Domain model to construct landmark sequences aligned with audio input.
  \item Introducing a novel feature fusion method block based on the KAN model.
  \item Paving the way for Audio-driven Talking face generation tasks based on landmark sequences.
\end{itemize}
% Đóng góp của chúng tôi sẽ được tóm tắt như sau:
% - Đề xuất mô hình KAN-Based Fusion of Dual Domain giúp xây dựng chuỗi landmarks khớp với giọng nói đầu vào.
% - Giới thiệu khối phương pháp hợp nhất đăc trưng mới dựa trên mô hình KAN.
% - Làm tiền cho các bài toán Audio-driven Talking face generation dựa trên chuỗi landmarks.

% Cấu trúc những chương tiếp theo của bài viết bao gồm 

\section{Related Work}
\label{sec:rework}
\subsection{Audio-driven Talking Face Generation}
The problem of audio-driven talking face generation aims to synthesize a target face that is compatible with the original audio sequence. The study of talking faces has been studied since the 90's \cite{lewis1991automated, guiard19963d, bregler2023video}. With advancements in computer technology and the widespread availability of network services, the applicability of this problem has expanded further. Some notable examples include applications in multimedia content production, video game creation \cite{xie2007coupled}, and dubbing for movies \cite{garrido2015vdub} or television programs \cite{charles2016virtual}. Additionally, it can also contribute to fields within community communication such as virtual assistants and virtual reality avatars. There have been many studies proposed to handle this problem, and in general we can divide these studies into two branches depending on how to process the input information.

% Bài toán về audio-driven talking face generation nhắm đến việt tổng hợp lên khuôn mặt mục tiêu tương thích với chuỗi audio ban đầu. Việc tạo khuôn mặt biết nói đã được nghiên cứu từ những năm 1990 [15–18]. Với sự tiến bộ của công nghệ máy tính và sự phổ biến của các dịch vụ mạng, tính ứng dụng của bài toàn này càng được mở rộng hơn. Một số ví dụ tiêu biểu có thể kể đến như ứng dụng để sản xuất các nội dung đa phương tiện, tạo trò chơi điện tử [19] và lồng tiếng cho phim [20] hoặc chương trình truyền hình [21]. Bên cạnh đó cũng có thể góp phần vào các lĩnh vực trong lĩnh vực giao tiếp cộng đồng như virtual assistant, virtual reality avatar. Đã có rất nhiều nghiên cứu được đề xuất để xử lý bài toán này, và nhìn chung chúng ta có thể chia các nghiên cứu này thành 2 nhánh tuỳ thuộc vào cách xử lý thông tin đầu vào.

\subsubsection{Image-Based Method:} This approach involves modifying one or several components based on the initial identity image. These methods primarily aim to create lip movements using a discriminator model \cite{chung2017you,zhou2019talking,kr2019towards,prajwal2020lip,cheng2022videoretalking,zhou2020makelttalk}. In one of the first studies, Chung et al. \cite{chung2017you} proposed generating lip-synced videos using an image-to-image translation approach. Then other outstanding studies were proposed such as Zhou et al. \cite{zhou2019talking} improved results by using disentangled audio-visual representation and recurrent neural networks. Prajwal et al. \cite{prajwal2020lip} proposed the Wav2Lip model, built on a GAN architecture, which employs a Lip-sync Discriminator to robustly between audio and video. However, these methods can only ensure synchronization between the generated mouth movements and audio, with minimal consideration for facial expressions or head movements. Zhou et al. \cite{zhou2020makelttalk} has developed a deep learning technique to create speaker-aware talking-head animations. This method uses an audio clip and a single image to predict facial landmarks based on disentangled audio content and the speaker's identity. The goal is to improve lip synchronization, personalize facial expressions, and capture natural head movements.

% Image-Based Method: Hướng tiếp cận này tập trung vào việc thay đổi một hoặc vài bộ phận dựa trên hình ảnh identity ban đầu. Chủ yếu những phương pháp này sẽ cố gắn tạo ra chuyển động môi with a discriminator model. \cite{kr2019towards, prajwal2020lip, cheng2022videoretalking}. Chung et al propose to generate lip-synced videos in an image-toimage translation manner, Zhou et al. [45] improve their results using disentangled audio-visual representation and recurrent neural networks. Chen et al. [9] leverage landmarks as intermediate representation and split the process into two stages. However, these methods can only promise the synchronization between generated mouths and audios. The results have barely any expression or head movements. Zhou et al. [47] successfully generate identity-related head movements, but their model also fails to control emotions.

\subsubsection{Video-Based Editing Methods:} 
Unlike the image-based method, this leverages the entire video frame to predict not only mouth movements, but also the facial structure, neck, and shoulders. This approach can provide a higher user experience compared to focusing solely on lip movements. However, the challenge with this type of view is its difficulty in performing well across all identities. Like Suwajanakorn et al. \cite{suwajanakorn2017synthesizing} accurately synthesized President Obama's face based on the audio of his speeches. His mechanism first selects the most suitable mouth region image from a database through audio-visual feature correlation and then integrates the selected mouth region with the original face. Ji et al. \cite{ji2021audio} addressed this challenge by introducing a method for predicting landmarks using Cross-Reconstructed Emotion Disentanglement, which separates speech into two distinct spaces: a time-independent emotional space and a time-dependent content space. Despite these breakthroughs, their results still have limitations, such as the robustness of landmarks being insufficient to effectively convey emotions through speech. Inspired by this approach and from \cite{ji2021audio} we will build a powerful model that can reconstruct the landmarks from audio in the most stable way to serve as a premise for the development of the audio-driven emotional talking face generation.

% Khác với hướng tiếp cận dựa trên hình ảnh, cách tiếp cận này sẽ tận dùng toàn khung hình video, để dự đoán không chỉ cuyển dộng của miệng mà còn của dáng mặt, phần cổ, và phần vai. Kết quả của hướng tiếp cận này sẽ đem lại trải nghiệm người dùng cao hơn so với việc chỉ tập trung vào chuyển động của môi. Tuy nhiên thử thách của hướng tiếp cận này là rất khó khăn để làm tốt với mọi ảnh identity. Suwajanakorn et al. [29] đã tổng hợp khuôn mặt của Tổng thống Obama rất chính xác, dựa trên âm thanh bài phát biểu của ông. Cơ chế này trước tiên là lấy hình ảnh vùng môi phù hợp nhất từ cơ sở dữ liệu thông qua tương quan đặc điểm nghe nhìn và sau đó kết hợp vùng môi được lấy với khuôn mặt ban đầu. Xinya el al. đã giải quyết khó khăn đó bằng việc giới thiệu phương pháp dự đoán lên các landmarks bằng kỹ thuật Cross-Reconstructed Emotion Disentanglement giúp phân tách lời nói thành hai không gian tách rời, tức là không gian cảm xúc không phụ thuộc vào thời lượng và không gian nội dung phụ thuộc vào thời lượng. Mặc dù đã có những bước đột phá tuy nhiên kết quả của họ vẫn còn một vài hạn chế như chất lượng landmarks chưa thật sự mạnh mẽ để có thể biểu đạt lại cảm xúc thông qua lời nói. Lấy cảm hứng từ hướng tiếp cận này và từ [] chúng tôi muốn xây dựng một mô hình mạnh mẽ có thể xây dựng chuỗi landmark một cách ổn định nhất để làm tiền đề cho sự phát triển của bài toán audio-driven emotional talking face generation.

\subsection{KAN: Kolmogorov-Arnold Networks}
\label{sec:kan}
Kolmogorov–Arnold Network (KAN) introduced in the paper \cite{liu2024kan}, is a model inspired by the Kolmogorov-Arnold representation theorem, KAN differentiates itself from the traditional Multi-Layer Perceptron (MLP) by replacing fixed activation functions with learnable functions, effectively eliminate the need about the linear weight matrix.
\subsubsection{Kolmogorov-Arnold Representation theorem}

Vladimir Arnold and Andrey Kolmogorov demonstrated that any continuous function \(f\) of \(n\) variables can be represented as a finite combination of continuous functions of a single variable and the binary operation of addition. Specifically, for any continuous function \(f(x_1, x_2, \ldots, x_n)\), there exist continuous functions \(\phi_i\) and \(\psi\) such that:
\begin{align}
\label{eq:kantheo}
f(x_1, x_2, \ldots, x_n) = \sum_{i=1}^{2n+1} \phi_i \left( \sum_{j=1}^{n} \psi_{ij}(x_j) \right)
\end{align}

% Vladimir Arnold và Andrey Kolmogorov đã chứng minh rằng bất kỳ hàm liên tục \(f\) nào của \(n\) biến đều có thể được biểu diễn dưới dạng tổ hợp hữu hạn của các hàm liên tục một biến và phép toán nhị phân của phép cộng. Cụ thể, đối với bất kỳ hàm liên tục \(f(x_1,x_2,...,x_n)\) luôn tồn tại các hàm liên tục \(phi_i\) và \(psi \) sao cho:

\subsubsection{The architecture of Kolmogorov-Arnold Networks} is based on an innovative concept where traditional weights are replaced by univariate functions at the network's boundaries. Instead of applying nonlinear transformations, each node in KAN simply aggregates the outputs of these functions. This approach contrasts with MLPs, which utilize linear transformations followed by nonlinear activation functions.
\begin{align}
\label{eq:kan}
KAN(x) = (\Phi_{L-1} \circ \Phi_{L-2} \circ ... \circ \Phi_{L_1} \circ \Phi_{L_0})x
\end{align}

% Kiến trúc của Kolmogorov-Arnold Networks (KAN) dựa trên một khái niệm đổi mới, trong đó các trọng số truyền thống được thay thế bằng các hàm đơn biến tại các biên của mạng. Thay vì áp dụng các biến đổi phi tuyến, mỗi nút trong KAN chỉ tổng hợp các đầu ra của các hàm này. Điều này khác với MLP, nơi sử dụng các biến đổi tuyến tính tiếp nối bởi các hàm kích hoạt phi tuyến.

In KAN's architecture, they used the Spline function as a learning mechanism. They serve as a replacement for the traditional weight parameters commonly found in networks. Thanks to the flexibility of this function, the model can adapt to correlations in the data by refining the shape to minimize loss. Due to that, the KAN network can adapt to complex data.

% Trong kiến trúc của KAN, họ đã sử dụng hàm Spline như một cơ chế học tập. Chúng đóng vai trò thay thế các tham số trọng lượng truyền thống thường thấy trong mạng. Nhờ vào tính linh hoạt của hàm số này giúp mô hình có thể thích ứng với các mối tương quan trong dữ liệu bằng các tinh chỉnh hình dạng để giảm thiểu mất mát. Nhờ vậy mạng KAN có thể thích ứng mới các dữ liệu phức tạp. 

\section{Proposed Method}
\label{sec:proposed}
\subsection{Overview}

\begin{figure}[t]
  \centering
  \includegraphics[width=\textwidth]{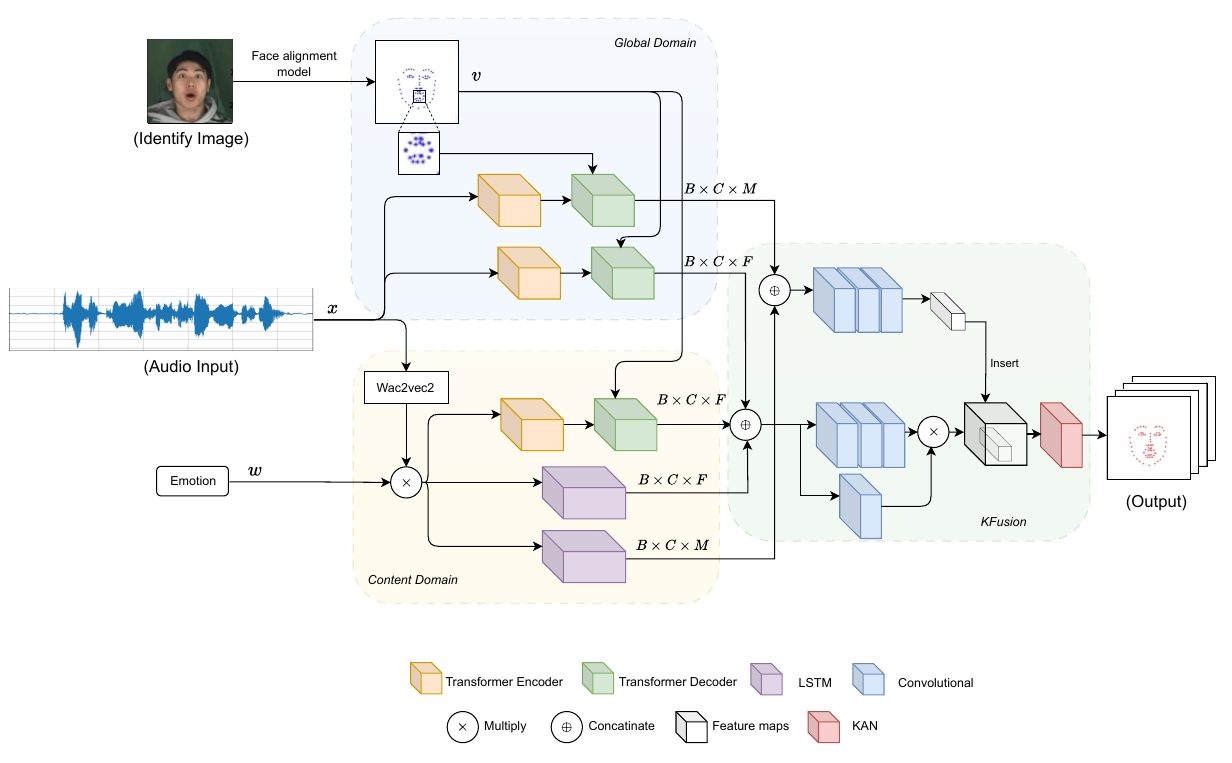}
  \caption{Overview of our model architecture, which consists of three distinct parts: Global Domain (blue background), Context Domain (yellow background), and KFusion (green background). The outputs of the two domains are features with dimensions \(B \times C \times F\) or \(B \times C \times M\), where \(F\) represents features for the entire face, and \(M\) represents features for the mouth region. The output of the entire model is a sequence of landmarks.}
  % Tổng quan kiến trúc mô hình của chung tôi, Trong đó có 3 phần riêng biệt, Global Domain (nền màu xanh dương), Context Domain (nền màu vàng), KFusion (Nền màu xanh lá cây). Outputs của hai domain là những đặc trưng có khích thước là B x C x F hoặc B x C x M. Trong đó F là đặc trưng dành cho toàn bộ khuôn mặt, M là đặc trưng cho phần khuôn miệng. Output của toàn bộ mô hình là sequence of landmarks.
  \label{fig:overview}
\end{figure}

Our model takes an audio signal and an identity image as input, and the output will be a sequence of landmarks corresponding to each frame of the ground truth video. As shown in the \cref{fig:overview}, we will process information in two separate domains. In the Global Domain, information will be directly extracted from the audio signal using a Transformer Encoder. Then combined with the identity landmarks of the entire face and mouth using a Transformer Decoder block. In the Context Domain branch, the signal will be transformed using the wav2vec2 \cite{baevski2020wav2vec} model and then multiplied by the emotional variable of the audio segment. We use a combination of LSTM and Transformer here to maximize the extraction of emotional information. We combine the information from the two domains using a KFusion block, where we correspondingly merge facial features and mouth features, and use KAN to adjust the landmarks to their most accurate positions.
% Mô hình của chúng tôi sẽ nhận đầu vào là một đoạn audio và một ảnh identity kết quả của mô hình là một chuỗi landmarks tương ứng với mỗi frame của video grouth truth. Như hình , chúng tôi sẽ xử lý thông tin theo 2 miền riêng biệt, trong đó Global domain sẽ trích xuất trực tiếp trên signal bằng Encoder Transformer. Sau đó kết hợp với landmark identity toàn bộ khuôn mặt và khuôn miệng bằng khối Decoder Transformer. Nhánh Context domain tín hiệu sẽ được biến đổi bằng mô hình Wav2vec2 sau đó nhân với biến cảm xúc của đoạn âm thanh. Chúng tôi sữ dụng kết hợp LSTM và Transformer ở đây để trích xuất tối đa thông tin về cảm xúc. Chúng tôi kết hợp thông tin của hai miền rong bằng khối KFusion, chúng tôi kết hợp tương ứng đặc trưng khuôn mặt, đặc trưng khuôn miệng tương ứng với nhau và dùng KAN để điều chỉnh landmark về vị trí đúng nhất.

\subsection{Global Domain}
\label{sec:global}
This block named is Global Domain because, in this block, features are extracted directly from the audio signal through the Transformer Encoder block \((\bf{E})\). These features then learn the correlation with the image landmarks. We are confident that using audio signals to learn facial movements is the correct approach because when speak, our facial movements directly reflect our expressions. For example, when we are angry, our voice exhibits strong variations, and correspondingly, our face moves up and down.

% Chúng tôi gọi phần này là Global Domain vì ở nhánh này sẽ trích xuất đặc trưng trực tiếp từ tín hiệu âm thanh thông qua khối Transformer Encoder. Sau đó cho đặc trưng này học tập tương quan với Landmark của hình ảnh Identify. Chúng tôi tin tưởng rằng sử dụng tín hiệu âm thanh để học chuyện động của khuôn mặt là việc làm đúng đắng vì khi nói chuyện, chúng ta phản xạ cử động mặt tương ứng. Lấy ví dụ, khi đang giận dữ âm thanh của chúng ta sẽ có dạng biến động mạnh mẽ, tương ứng với đó khuôn mặt cũng sẽ cữ động lên xuống.
By using the Transformer Decoder block \((\bf{D})\) and the landmarks from the identity image \(v\) as a guiding variable, the audio features are gradually structured into facial landmark features through Multi-Head Attention blocks. These blocks create masks and then apply features from the audio signal \(x\), thus shaping the landmarks. A Convolutional block \((\bf{Conv})\) is used immediately after the Decoder to transform the features \(x\) to be suitable for multiplication with the features \(v\) within the Decoder block. Additionally, we isolate the mouth landmarks and apply similar methods to extract features for mouth movements. The output of this stage will be \(gb \in \mathbb{R}^{B \times C \times F}\) and \(gb_m \in \mathbb{R}^{B \times C \times M}\). At this stage, the process can be formulated as follows:

% Bằng cách sử dụng khối Transformer Decoder và landmark từ hình ảnh Identity \(v\) như một biến hướng dẫn. Đặc trưng dạng âm thanh sẽ dần được kiến trúc thành đặc trưng landmark của khuôn mặt nhờ các khối Multi Head Attention. Khối này sẽ xây dựng các mask sau đó apply đặc trưng từ tín hiệu âm thanh \(x\), qua đó kiến trúc lên hình dạng của landmark. Một khối Convolutional sẽ được sử dụng ngay sau khối Decoder với mục đích biến đổi đặc trưng \(x\) để phù hợp khi nhân với đặc trưng \(v\) bên trong khối Decoder. Bên cạnh đó chúng tôi lấy riêng phần landmark của khuôn miệng và áp dụng các tương tự để trích xuất đặc trưng cho sự chuyển động của khuôn miệng. Output của giai đoạn này sẽ có dạng \(B \times C \times F\) và \(B \times C \times M\). Có thể viêc quá trình ở giai đoạn này dứoi dạng công thức như sau.

\begin{align}
\label{eq:global}
L_m &= \{ l_{i_{1}}, l_{i_{2}}, ..., l_{i_{M}} \}\\
v_m &= v[:,:,L_m]  \\
gb &= D_f(Conv(E(x), v) \\
gb_m &= D_m(Conv(E(x), v_m) 
\end{align}

Here, \(L_m\) is the set of positions of mouth points on the landmarks map. The result of this region is two features: \(gb\) is the feature across the entire face, and \(gb_m\) is the feature of the mouth area.

% Trong đó L_m là tập hợp các vị trí của khuông miệng trên bảng đồ landmarks. kết quả của miền này là 2 đặc trưng gb là đặc trưng trên toàn bộ khuôn mặt, gbm là đặc trưng của vùng miệng.

\subsection{Content Domain}
\label{sec:content}
In this domain, the input is an audio signal, and there will be 3 outputs. Among them \(ct,f \in \mathbb{R}^{B \times C \times F}\) và \(m \in \mathbb{R}^{B \times C \times M}\). The audio \((x)\) will first be transformed into representative vectors for the script using the wav2vec2 model \cite{baevski2020wav2vec} to be \((x')\). These vectors will then be multiplied by the emotion variable \((w)\), thereby amplifying the information similarities between speech and emotion.

% Ở domain này, input là tín hiệu âm thanh và sẽ có 3 output. Trong đó \(ct,f in B \times C \times F \) và \(c in B \times C \times M\)
% trước tiên âm thanh sẽ được biến đổi bằng mô hình wav2vec2 \cite{} thành các vector đại diện cho script. Vector này sẽ được nhân với biến cảm xúc qua đó sẽ khuếch đại những thông tin tương đồng giữa lời nói và cảm xúc. 
\begin{align}
\label{eq:wav2vec}
x' = wav2vec2(x) \times w
\end{align}

Then, the information will be processed through three branches. One branch will be similar to the branch in the Global Domain. However, the difference here is that the input information of the Encoder block is the script representative vector \((x')\), thereby learning a deeper layer of information. Additionally, we use two LSTM networks to extract mouth and face information from the vector \((x')\).
% Sau đó, thông tin sẽ được học tập theo 3 nhánh. Một nhánh sẽ tương tự cách nhánh trong miền global. Tuy nhiên, khác biệt ở đây nằm ở việc thông tin đầu vào của khối Encoder là vector đại diện script, qua đó một lớp thông tin sâu hơn sẽ được học tập. Bên cạnh đó chúng tôi cũng dùng 2 mạng LSTM để trích xuất thông tin miệng và mặt từ vector \(x'\)

\subsection{KFusion: KAN-based Fusion}
The KFusion block receives the outputs from the Global Domain and the Content Domain. These outputs have two different sizes. This block will extract information from each corresponding size and will return the predicted landmarks result. We concatenate the outputs with the corresponding sizes together as in the formula \cref{eq:concatinate}
% Đây là phần trọng tâm của chúng tôi. Khối KFusion nhận các output của Global domain và Content domain. Các output đó có 2 kích thước khác nhau. Khối này sẽ trích xuất thông tin với mỗi loại kích thước tương ứng và sẽ trả về kết quả landmarks dự đoán \((y')\).
% Chúng tôi tiến hành concatinate các output với các kích thước tương ứng lại với nhau như công thức \cref{eq:concatinate}.

\begin{align}
\label{eq:concatinate}
x_m = m \oplus gb_m \\
x_f = f \oplus gb \oplus ct
\end{align}
For each obtained feature map value, we pass it through a block consisting of \(Conv(k=1) \rightarrow  Conv(k=3) \rightarrow  Conv(k=1)\) (k is kernel size). Inspired by the Bottleneck Block from \cite{he2016deep}, we denote this block as \(B\). Thus, we have \(x'm = B(x_m) \). For the feature representation of the entire face, we add a Residual branch since the whole face information is crucial in this part. The formula for \(x'_f\) is shown in \cref{eq:face}. 

% Với mỗi giá feature maps vừa thu được, chúng tôi sẽ cho qua một khối gồm Conv(k=1) -> Conv(k=3) -> Conv(k=1). Được lấy ý tưởng theo Bottleneck Block từ \cite{he2016deep}, chúng tôi đặt khối này là \(B\). Qua đó ta có \(x'm = B(x_m) \). Đối với đặc trưng biểu diễn toàn khuôn mặt, chúng tôi cho thêm một nhánh Residual vì thông tin toàn khuôn mặt rất trong trọng trong phần này. Công thức \(x'_f\) được hiển thị tại \cref{eq:face}

Next, we insert \(x'_m\) into \(x'_f\) at the corresponding positions \(L_m\), as shown in \cref{fig:overview}. The formula for this step is presented in \cref{eq:face2}. Finally, we use the KAN network to predict the landmarks \(y'\). The KAN model has been described in \cref{sec:kan}. The entire process of this fusion stage is presented below.

% Tiếp theo, chúng tôi chèn x'_m vào x'_f theo các vị trí L_m tương ứng. Như hình \cref{static/modeloverview.pdf}. Công thức của bước này được trình bày như \cref{eq:face2}. Cuối cùng, chúng tôi sử dụng mạng KAN để đự đoán lên các landmarks. Mô hình KAN đã được chúng tôi trình bày tại \cref{2.2}. Toàn bộ quá trình của giai đoạn hợp nhất này được trình bày ngay bên dưới.

\begin{align}
\label{eq:face}
& x'_f = B(x_f) \otimes Conv(x_f) \\
\label{eq:face2}
& x'_f[:,:,L_m] \leftarrow  x'_m \\
\label{eq:face3}
& y' =  KAN(x'_f)
\end{align}

\section{Experiments}
\label{sec:exp}

\subsection{Implementation Details}
\subsubsection{Dataset.} Throughout the entire training, These use the MEAD dataset \cite{wang2020mead}. The testing phase was run on 2 sets MEAD and CREMA-D. We apply several augmentation methods and use the Facial alignment framework to extract landmarks from the videos \cite{bulat2017far}. The landmarks that have been extracted include a total of 68 points.

\begin{figure}[tb]
  \centering
  \includegraphics[width=\textwidth]{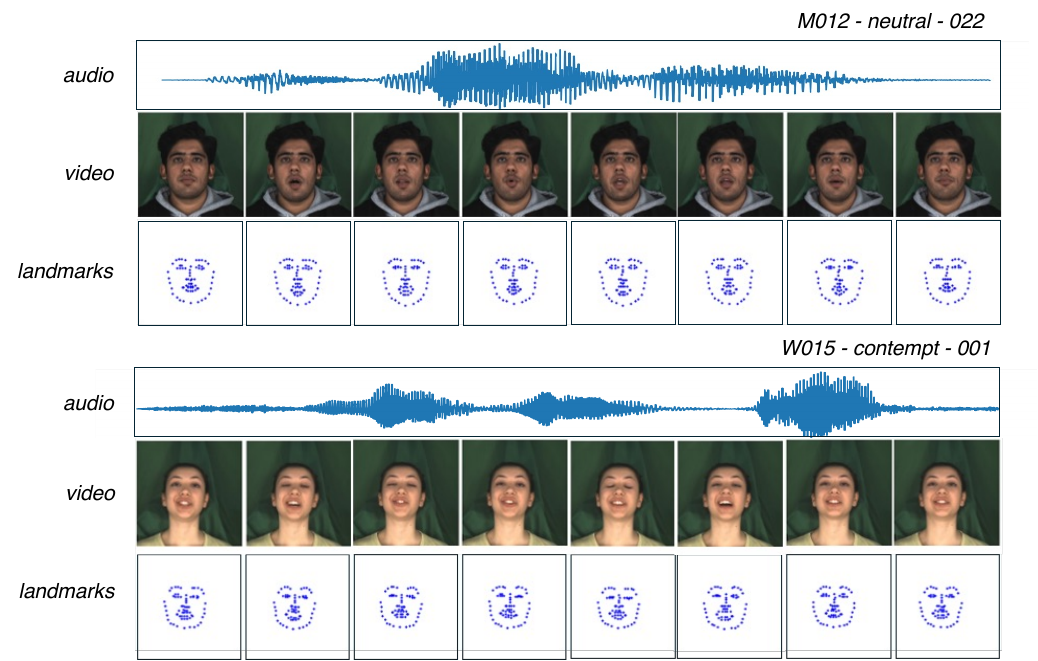}
  \caption{MEAD: The dataset we use includes video and accompanying audio. ``M'' denotes male and ``W'' denotes female. There are a total of 8 emotions in the conversations. Additionally, we also extract landmarks from the video to serve as ground truth for our problem.
  }
  % MEAD: Tập dữ liệu mà chúng tôi sử dụng bao gồm video và âm thanh đi kèm. Tập dữ liệu này rất đa dạng về giới tính, sắc tộc. Trong đó M kí hiệu là nam và W kí hiệu cho nữ. Có tổng cộng 8 cảm xúc khi nói chuyện. Ngoài ra chúng tôi cũng extract landmarks từ video để làm grouth truth cho bài toán của chúng tôi.
  \label{fig:sample}
\end{figure}

% Trong toàn bộ quá trình huấn luyện và kiểm tra, chúng tôi sử dụng bộ dataset MEAD \cite{wang}. Chúng tôi có sử dụng một vài phương pháp biến đổi, và dùng framework Facial alignment để trích xuất landmarks từ video.
MEAD is very large with nearly 40 hours of audio-visual clips recorded for each person and view. The participants in this dataset include 60 people of different races and genders. Each person exhibits 8 emotions, 3 levels of intensity, and 7 different view angles. However, only the frontal angle was used for this experiment. Each video has a resolution of \(1920 \times 1080 \times 3\), which is very large, then was resized and cropped to \(256 \times 256 \times 3\) and extracted landmarks from this resized image. The original sound in the dataset is in stereo audio format with a sample rate of 48000 hertz, which has been converted to mono audio format and the sample rate to 16000 hertz. \cref{fig:sample} shows some samples from the MEAD dataset have transformed. The audio is the output of our task, and the landmarks are what we aim to achieve. The video length is set to 1s with an FPS of 30. 

% Tập dữ liệu MEAD là một tập dữ liệu rất lớn với gần 40 hours of audio-visual clips are recorded for each person and view. Đối tượng tham gia trong tập dữ liệu này là 60 person với chủng tộc, giới tính khác nhau. Với mỗi người sẽ có 8 cảm xúc, 3 intensity of level and 7 different view angles. Tuy nhiên chúng tôi chỉ sử dụng góc trực tiếp cho bài thí nghiệm này. Mỗi video có kích thước là \(1920 \times 1080 \times 3\) kích thước này là rất lớn,  nên đã được đã thay đổi kích thước và cắt kích thước còn \(256 \times 256 \times 3\) và extract landmarks trên hình ảnh mới này. Âm thanh gốc trong tập dữ liệu là dạng stereo audio với sample rate là 48000 hertz được biến đổi thành dạng môn audio và sample rate thành 16000 hertz. Hình 3 là giới thiệu một vài mẫu từ tập dữ liệu MEAD mà chúng tôi đã biến đổi. Trong đó audio chính là output của bài toán, landmarks là thứ mà chúng tôi cần đạt được.

 CREMA-D dataset was used in the testing phase \cite{cao2014crema}, which was collected from 48 actors and 43 actresses, totaling 7442 samples. The emotion labels of this dataset encompass 6 different emotions, all of which are also present in the MEAD dataset. The same preprocessing methods were applied, reconstructing the input to be suitable for the model.

% Bên cạnh đó, một bộ dataset khác được sử dụng là CREMA-D được dùng để đánh giá. Bộ dataset được thu thập từ 48 actor and 43 actress tổng cộng là 7442 sample. Nhãn cảm xúc của bộ này có 6 cảm xúc khác nhau, tất cả những cảm xúc này điều có trong tập MEAD. Chúng tôi sử lý dữ liệu bằng cách tương tự như tập MEAD.

\subsubsection{Experimental setup.}

The experiments were run with the number of epochs set to 200. The loss function used is the Mean Square Error. The optimization function used in the experiments is Adam with an initial learning rate of \(3e-4\). The learning rate schedule used is Cosine Annealing with the maximum number of iterations equal to the number of epochs, meaning the learning rate will be adjusted following a sine wave pattern with a minimum learning rate of \(1e-6\). The hardware we used is an NVIDIA GeForce 4090 with 24GB GPU memory.
% Các thực nghiệm được đã được chạy với số lượng epochs được thiết lập là 200. Hàm tối ưu được dùng trong bài thực nghiệm là Adam với init learning rate là \(3e-4\). Learning rate schedule được sử dụng là Cosine Annealing với maximum number of iterations bằng với số lượng epochs đồng nghĩa learning rate sẽ được gia giảm theo đồ thị hình sine và minimun learning rate là \(1e-6\) . Phần cứng chúng tôi sử dụng là NVIDIA GeForce 4090 với 24GB GPU bộ nhớ.

\subsubsection{Comparing Methods.}
Evaluation metrics are based on two metrics: Landmark distance (LD), and Landmark velocity difference (LVD). These two metrics were introduced in the paper \cite{zhou2020makelttalk}. LD: The average value of the Euclidean distance between all predicted facial landmark locations \cref{eq:ld}. LVD: The average Euclidean distance between the reference datum velocity and the predicted velocity \cref{eq:lvd}.

% Các phương pháp được dùng để so sánh bao gồm \cite{}. Trong đó chúng tôi sẽ đánh giá dựa trên 2 thang đo là Landmark distance (LD), và Landmark velocity difference (LVD). Hai thang đo này được giới thiệu trong bài \cite{}. LD: The average value of the Euclidean distance between all predicted facial landmark locations. LVD:  he average difference in speed (velocity) between predicted landmark locations and their actual counterparts in a video. This difference is then normalized by the width of the face to account for size variations. By analyzing these velocity discrepancies across all facial landmarks, D-V provides valuable insights into the model's ability to capture the dynamics of facial motion.

\begin{align}
\label{eq:ld}
&v_{x,i} = x_{i,t} - x_{i,t-1} \\
LD &= \frac{1}{N} \sum_{i=1}^N \sqrt{(x_i - \hat{x}_i)^2 + (y_i - \hat{y}_i)^2} \\
\label{eq:lvd}
LVD &= \frac{1}{N} \sum_{i=1}^N \sqrt{(v_{x,i} - \hat{v}_{x,i})^2 + (v_{y,i} - \hat{v}_{y,i})^2}
\end{align}

In there \((x, y)\) represent for reference landmark in $i$ position,
\((\hat{x}, \hat{y})\) represent for predicted landmark and $t$ represent for the frame.

The process will be performed on two components of the result: the entire face and the mouth segment marked as landmarks from positions 48-67. Therefore, there will be a total of 4 evaluation methods: M-LD, M-LVD, F-LD and F-LVD. The models used for comparison are outstanding models introduced from 2019 to the present, including \cite{chen2019hierarchical,wang2020mead,song2022everybody,ji2021audio,sinha2022emotion, cao2023speechsyncnet}.
% Quá trình sẽ thực hiện trên 2 thành phần của kết quả là toàn bộ khuôn mặt và phân vùng miệng được đánh dấu là các landmark từ vị trí số 48-67. Vì thế sẽ có tất cả là 4 phương pháp đánh giá là M-LD, M-LVD, F-LD và F-LVD. Các mô hình dùng để so sánh là những mô hình nổi bật được giới thiệu từ năm 2019 đến nay bao gồm \cite{}

\subsection{Experimental Results}
\subsubsection{Qualitative Comparisons}
\begin{figure}[p]
  \centering
  \includegraphics[width=\textwidth]{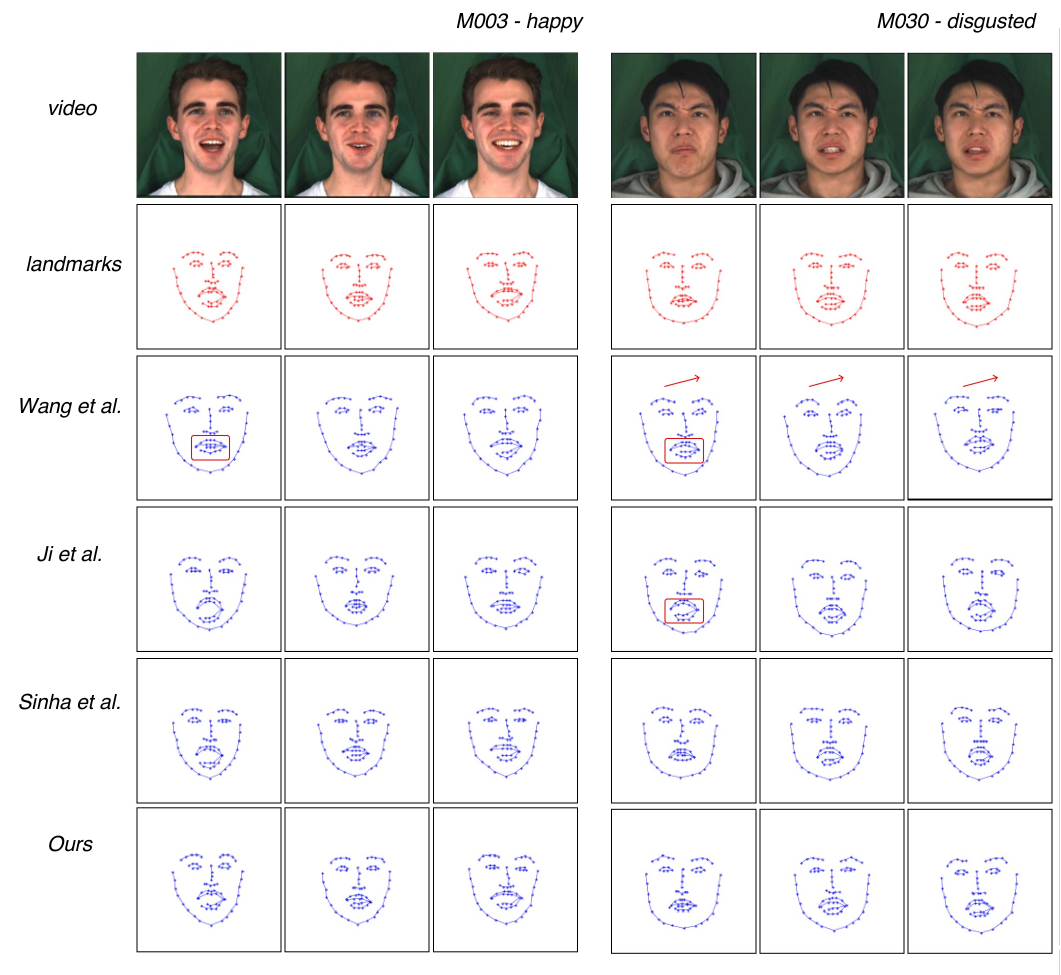}
  \caption{Quantitative comparison with other methods on the MEAD dataset for samples M003 and M030. The red landmarks are the target. The last row shows the results of the proposed method. In the third row, the model by Wang et al \cite{wang2020mead}. has red squares indicating inaccuracies in mouth movements and red arrows pointing to incorrect facial orientations. The fourth row shows similar issues to the model by Ji et al \cite{ji2021audio}.
  }
  % So sánh định lượng với các phương pháp khác trên tập dữ liệu MEAD mẫu M003 và mẫu M030. Trong đó landmarks màu đỏ đó là target. Hàng cuối cùng là kết quả của phương pháp đề xuất. Hàng thứ 3 mô hình của Wang et al. có ô vuông màu đỏ biểu thị sự sai lệch trong cử động miệng, dấu mũi tên đỏ chỉ hướng mặt bị sai lệch. Hàng thứ 4 mô hình của Ji et al. tương tự.
  \label{fig:quality}
\end{figure}

The results of the qualitative comparison are shown in the \cref{fig:quality}. The proposed method has been able to construct facial landmarks as expected. The movements of the mouth and facial corners are not too different from the target landmarks. Comparison with other models shows the prominence of incorrect mouth movements in the model \cite{wang2020mead}, \cite{kr2019towards} , as seen in rows 3 and 4.
% Kết quả so sánh định tính được hiển thị tại hình, Phương pháp đề xuất đã có thể xây dựng được landmark của khuôn mặt khá tốt. Các cử động của miệng và góc mặt không quá sai lệch so với landmarks target. So sánh với các mô hình khác cho thấy sự nổi khi cử động vùng miệng của mô hình \cite{} không đúng xem tại hàng thứ 3, 4.

\subsubsection{Quantitative Comparisons}
For the MEAD dataset, the proposed model has shown its effectiveness when evaluated quantitatively. According to \cref{tab:compair}, the proposed model achieved the best results in terms of M-LD and F-LD values compared to other models. The difference from the second-best model is approximately $\approx0.6$ in M-LD value and $\approx0.02$ in F-LD value. For M-LVD and F-LVD values, the lowest values were achieved by Sinha et al. \cite{sinha2022emotion}. However, the proposed model is also competitive, with the difference not being significant, around $\approx0.04$ in M-LVD value and $\approx0.09$ in F-LVD.
With CREMA-D, model results still achieves the optimal value on M-LD and F-LD. For M-LVD and F-LVD, some remarkable results were also obtained that were higher than the method of Vougioukas et al. \cite{vougioukas2020realistic} around $\approx0.2$.
% Đối với M-LVD và F-LVD cũng thu được một số kết quả đáng ghi nhận cao hơn phương pháp của Vougioukas et al. \cite{} khoảng 0.2
% Mô hình đề xuất đã cho thấy sự hiệu quả khi đánh giá định lương. Theo \cref{}, mô hình đề xuất đã đạt kết quả tối ưu nhất nhất trên giá trị M-LD và F-LD khi so với các mô hình còn lại. Độ chênh lệch so với mô hình tối ưu thứ hai là khoảng ~0.6 trên giá trị M-LD và khoảng ~0.02 trên giá trị F-LD. Đối với giá trị M-LVD và F-LVD mô hình đạt giá trị thấp nhất là Sinha et al. \cref{}. Tuy nhiên, mô hình đề xuất cũng không kém cạnh khi độ chênh lệch là không quá lớn khoảng ~0.04 trên giá trị M-LVD và 0.09 trên F-LVD.

% So sánh định lượng mô hình của chúng tôi và các mô hình trước đây. Tất cả điều được đánh giá trên tập dữ liệ MEAD. Các phương pháp đánh giá bao gồm M-LD, M-LVD, F-LD, F-LVD. Trong đó M là kí hiệu cho các landmarks vùng miệng, F là kí hiệu của toàn bộ landmarks. Tất cả các phương thức đánh giá định nghĩa giá trị càng bé là càng tốt.

\begin{table}[tb]
  \caption{Quantitative comparison of our model and previous models. There are two datasets used MEAN and CREMA-D. The evaluation methods include M-LD, M-LVD, F-LD, and F-LVD. In these, M denotes the mouth region landmarks, and F denotes all landmarks. In all evaluation methods, a smaller value indicates better performance. Bold print represents the most optimal value, and underline represents the second optimal value. The lower the value of the evaluation method is the better.
  }
\label{tab:compair}
\centering
  \begin{tabular}{@{}l|c|c|c|c|c@{}}
    \toprule
    \multicolumn{6}{c}{\bf{MEAD}\cite{wang2020mead}} \\
    \midrule
\bf{Method} & \bf{   Year   } & \bf{  M-LD  (\(\downarrow\))  } & \bf{  M-LVD (\(\downarrow\))  }             & \bf{  F-LD  (\(\downarrow\))  } & \bf{  F-LVD (\(\downarrow\))  } \\ 
\midrule
Chen et al. \cite{chen2019hierarchical}  & {2019} & {3.27} & {2.09} & {3.82} & 1.71       \\ 
Wang et al. \cite{wang2020mead}          & {2020} & {2.52} & {2.28} & {3.16} & 2.01       \\ 
Song et al. \cite{song2022everybody}     & {2020} & {2.54} & {1.99} & {3.49} & 1.76       \\ 
Ji et al. \cite{ji2021audio}             & {2021} & {2.45} & {1.78} & {3.01} & 1.56       \\ 
Sinha et al. \cite{sinha2022emotion}   & {2022} & {2.18} & \bf{0.77} & \underline{1.24} & \bf{0.50}  \\ 
Cao et al. \cite{cao2023speechsyncnet}   & {2023} & \underline{1.92}   & {0.97}      & {1.25}   & {0.70} \\ 
\midrule
\bf{Ours}                                &        & \bf{1.33} & \underline{0.81} & \bf{1.22} & \underline{0.59} \\ 
\toprule
\multicolumn{6}{c}{\bf{CREMA-D}\cite{cao2014crema}} \\
\midrule
Vougioukas et al.  \cite{vougioukas2020realistic} & 2020 & 2.90 & \bf{0.42}  & 2.80 & \bf{0.34} \\
Eskimez et al.     \cite{eskimez2021speech}       & 2021 & 6.14 & \underline{0.49} & 5.89 & \underline{0.40}  \\
Sinha et al.       \cite{sinha2022emotion}        & 2022 & \underline{2.41} & 0.69 & \underline{1.35} & 0.46 \\
\midrule
\bf{Ours} &                                       & \bf{2.37} & 0.66           & \bf{1.23} & 0.52 \\         
\bottomrule
\end{tabular}
\end{table}

\subsubsection{Ablation Studies}
We conducted several ablation studies to demonstrate the effectiveness of each component within the model. There were four experiments as described in \cref{tab:ablation}. The experiment without the KFusion block (w/o KFusion) yielded the worst results, with values across the entire face (F) being approximately $\approx 8.6$ for LD and $\approx 10$ for LVD, indicating the effectiveness of the KFusion block. Additionally, we tried replacing the KAN network with an MLP network (repl MLP). The results were acceptable, with the M-LD value nearly optimal, but the other values remained quite high. We also conducted experiments by disabling one of the branches to observe their effectiveness. When we did not use the LSTM layers in the Context Domain \cref{sec:content} (w/o LSTM) or did not use the Global Domain \cref{sec:global} (w/o Global), the results were quite similar; however, using LSTM in the Content Domain yielded more optimal results. This demonstrates that the Content Domain block has learned very effectively, especially for the LD value, which is the average Euclidean distance between the corresponding predicted and reference positions.
% Chúng tôi đã thực nghiệm một số ablation studies để chứng mình độ hiểu quả của từng thành phần bên trong mô hình. Trong đó có 4 thí nghiệm như \cref{}. Với thí nghiệm không dùng khối KFusion, đây là kết quả tệ nhất. Trong đó các giá trị trên toàn bộ khuôn mặt (F) khoảng $\approx 8.6$ đối với LD và $\approx 10$ đối với LVD cho thấy độ hiệu quả của khối KFusion mang lại. Bên cạnh đó chúng tôi còn thử thay thế mạng KAN bằng mạng MLP. Kết quả là tạp chấp nhận được khi giá trị M-LD đạt gần tối ưu nhất nhưng những giá trị còn lại vẫn còn rất cao. Ngoài ra chúng tôi cùng tiến hành tắt một trong các nhánh để xem độ hiệu quả mà chúng đem lại. Khi không sử dụng các lớp LSTM trong Context Domain \cref{} hoặc không sử dụng khối Global Domain \Kết quả khá tương đồng tuy nhiên nếu sử dụng LSTM trong Content Domain sẽ cho ra kết quả tối ưu hơn. Điều này chúng minh rằng khối Content Domain đã học tập rất hiệu quả đặc biệt là đối với giá trị LD, trung bình Euclidean các vị trí tương ứng giữa dự đoán và reference.
\begin{table}[tb]
\centering
\caption{Comparison of results between ablation study experiments. In these experiments, we try disabling one of the domains or replacing the KFusion block with an MLP. The bold print represents the most optimal value, and the underline represents the second optimal value. The lower the value of the evaluation method the better.}
% So sánh kết quả giữa các thí nghiệm ablation studie. Trong đó chúng tôi thử disable một trong các domain hoặc thay thế khối KFusion thành Linear.
\label{tab:ablation}
\begin{tabular}{@{}|l|c|c|c|c|@{}}
\toprule
\bf{Method} & \bf{  M-LD  (\(\downarrow\))  } & \bf{  M-LVD (\(\downarrow\))  } & \bf{  F-LD  (\(\downarrow\))  } & \bf{  F-LVD } (\(\downarrow\)) \\
\midrule
w/o KFusion            & {3.300}      & {2.451}       & {8.624}      & {10.00}       \\
w/o LSTM              & {2.799}      & {1.787}       & {5.490}      & {6.123}       \\
w/o Global            & {2.120}      & {1.373}       & {4.126}      & {6.015}       \\
repl MLP                & {1.380}      & {2.520}       & {3.990}      & {3.480}       \\
\midrule
\bf{Proposed}         & \bf{1.333}   & \bf{0.808}    & \bf{1.225}   & \bf{0.599}  \\
\bottomrule
\end{tabular}
\end{table}

\section{Conclusion}
In this paper, we introduce a model that generates a sequence of landmarks from an input consisting of audio and an identity image. The proposed model is named \textit{``KFusion of Dual-Domain''} and is divided into three components. The Global Domain block learns information directly from the audio signal, which helps in better learning of facial movements. In parallel, the Content Domain block learns information based on emotion vectors extracted from the wav2vec2 model, which helps in better reconstruction of facial expressions by learning emotional information. Finally, the KFusion block synthesizes information from the two aforementioned domains. It integrates features based on their respective dimensions, including the mouth region and the entire face. A KAN network is used at the end to predict the landmarks. Ablation experiments have demonstrated the value of each component and the effectiveness of their combination.
% Trong bài viết này chúng tôi giới thiệu mô hình generate ra sequence of landmarks từ input là audio và một hình ảnh Identity. Mô hình đề xuất có tên là KFusion of Dual-Domain trong đó được chia thành 3 thành phần. Khối Global Domain học thông tin trực tiếp từ âm thanh, điều này cho sẽ giúp học tập các chuyển động trên khuôn mặt một cách tốt hơn. Song song với đó, khối Content Domain sẽ học thông tin dựa trên vector cảm xúc được trích xuất từ mô hình wac2vec2, khối này sẽ học tập các thông tin của cảm xúc giúp cho việc tái tạo các biểu cảm được tốt hơn. Cuối cùng khối KFusion sẽ tổng hợp các thông tin từ 2 Domain trên. Nó tổng hợp đặc trưng theo từ kích thước tương ứng gồm có vùng miệng và toàn bộ khuôn mặt. Một mạng KAN được dùng ở cuối cùng để dự đoán các lên các điểm landmarks. Thí nghiệm ablation đã chứng minh được giá trị của từng phần và khi kết hợp các thành phần lại với nhau.
The qualitative results shown in the figure demonstrate that the proposed model has performed as expected and has improved compared to previous models. The quantitative results also indicate the same. The proposed model has achieved remarkable results, particularly on F-LD and M-LD, when evaluated on two datasets, MEAD and CREMA-D.
% Kết quả định tính được hiển thị tại hình cho thấy rằng mô hình đề xuất đã hoạt động đúng như mong muốn và đã có cải thiện hơn so với các mô hình trước đây. Kết quả định lượng cũng cho thấy điều tương tự. Mô hình đề xuất có kết quả rất đáng ghi nhận đặt biệt là trên F-LD và M-LD khi đánh giá trên 2 bộ dataset là MEAD và CREMA-D.
\subsubsection{Limitation}
Our paper presents a method to generate landmarks from audio, which serves as a foundation for generating faces from audio. The architecture of the proposed model is still simple. In the future, we will continue to develop it to address the larger problem of Audio-Driven Talking Face Generation.
% Bài viết của chúng tôi trình bày phương pháp generate landmarks từ audio, đây là nền tảng cho các bài toán generate khuôn mặt từ âm thanh. Kiến trúc mô hình đề xuất còn đơn giản. Trong tương lai, chúng tôi sẽ tiếp tục phát triển để giải quyết bài toán lớn hơn là Audio-Driven Talking Face Generation.

\label{sec:conc}

% --------------------CONTENT-------------------------

\bibliographystyle{splncs04}
\bibliography{main}

\begin{thebibliography}{10}
\providecommand{\url}[1]{\texttt{#1}}
\providecommand{\urlprefix}{URL }
\providecommand{\doi}[1]{https://doi.org/#1}

\bibitem{baevski2020wav2vec}
Baevski, A., Zhou, Y., Mohamed, A., Auli, M.: wav2vec 2.0: A framework for self-supervised learning of speech representations. Advances in neural information processing systems  \textbf{33},  12449--12460 (2020)

\bibitem{bregler2023video}
Bregler, C., Covell, M., Slaney, M.: Video rewrite: Driving visual speech with audio. In: Seminal Graphics Papers: Pushing the Boundaries, Volume 2, pp. 715--722 (2023)

\bibitem{bulat2017far}
Bulat, A., Tzimiropoulos, G.: How far are we from solving the 2d \& 3d face alignment problem?(and a dataset of 230,000 3d facial landmarks). In: Proceedings of the IEEE international conference on computer vision. pp. 1021--1030 (2017)

\bibitem{cao2014crema}
Cao, H., Cooper, D.G., Keutmann, M.K., Gur, R.C., Nenkova, A., Verma, R.: Crema-d: Crowd-sourced emotional multimodal actors dataset. IEEE transactions on affective computing  \textbf{5}(4),  377--390 (2014)

\bibitem{cao2023speechsyncnet}
Cao, X.N., Trinh, Q.H., Ho, V.S., Tran, M.T.: Speechsyncnet: Speech to talking landmark via the fusion of prior frame landmark and the audio. In: 2023 IEEE International Conference on Visual Communications and Image Processing (VCIP). pp.~1--5. IEEE (2023)

\bibitem{charles2016virtual}
Charles, J., Magee, D., Hogg, D.: Virtual immortality: Reanimating characters from tv shows. In: European Conference on Computer Vision. pp. 879--886. Springer (2016)

\bibitem{chen2019hierarchical}
Chen, L., Maddox, R.K., Duan, Z., Xu, C.: Hierarchical cross-modal talking face generation with dynamic pixel-wise loss. In: Proceedings of the IEEE/CVF conference on computer vision and pattern recognition. pp. 7832--7841 (2019)

\bibitem{cheng2022videoretalking}
Cheng, K., Cun, X., Zhang, Y., Xia, M., Yin, F., Zhu, M., Wang, X., Wang, J., Wang, N.: Videoretalking: Audio-based lip synchronization for talking head video editing in the wild. In: SIGGRAPH Asia 2022 Conference Papers. pp.~1--9 (2022)

\bibitem{chung2017you}
Chung, J.S., Jamaludin, A., Zisserman, A.: You said that? arXiv preprint arXiv:1705.02966  (2017)

\bibitem{eskimez2021speech}
Eskimez, S.E., Zhang, Y., Duan, Z.: Speech driven talking face generation from a single image and an emotion condition. IEEE Transactions on Multimedia  \textbf{24},  3480--3490 (2021)

\bibitem{garrido2015vdub}
Garrido, P., Valgaerts, L., Sarmadi, H., Steiner, I., Varanasi, K., Perez, P., Theobalt, C.: Vdub: Modifying face video of actors for plausible visual alignment to a dubbed audio track. In: Computer graphics forum. vol.~34, pp. 193--204. Wiley Online Library (2015)

\bibitem{guiard19963d}
Guiard-Marigny, T., Tsingos, N., Adjoudani, A., Benoit, C., Gascuel, M.P.: 3d models of the lips for realistic speech animation. In: Proceedings Computer Animation'96. pp. 80--89. IEEE (1996)

\bibitem{he2016deep}
He, K., Zhang, X., Ren, S., Sun, J.: Deep residual learning for image recognition. In: Proceedings of the IEEE conference on computer vision and pattern recognition. pp. 770--778 (2016)

\bibitem{ji2021audio}
Ji, X., Zhou, H., Wang, K., Wu, W., Loy, C.C., Cao, X., Xu, F.: Audio-driven emotional video portraits. In: Proceedings of the IEEE/CVF conference on computer vision and pattern recognition. pp. 14080--14089 (2021)

\bibitem{kr2019towards}
KR, P., Mukhopadhyay, R., Philip, J., Jha, A., Namboodiri, V., Jawahar, C.: Towards automatic face-to-face translation. In: Proceedings of the 27th ACM international conference on multimedia. pp. 1428--1436 (2019)

\bibitem{lewis1991automated}
Lewis, J.: Automated lip-sync: Background and techniques. The Journal of Visualization and Computer Animation  \textbf{2}(4),  118--122 (1991)

\bibitem{liu2024kan}
Liu, Z., Wang, Y., Vaidya, S., Ruehle, F., Halverson, J., Solja{\v{c}}i{\'c}, M., Hou, T.Y., Tegmark, M.: Kan: Kolmogorov-arnold networks. arXiv preprint arXiv:2404.19756  (2024)

\bibitem{pataranutaporn2021ai}
Pataranutaporn, P., Danry, V., Leong, J., Punpongsanon, P., Novy, D., Maes, P., Sra, M.: Ai-generated characters for supporting personalized learning and well-being. Nature Machine Intelligence  \textbf{3}(12),  1013--1022 (2021)

\bibitem{prajwal2020lip}
Prajwal, K., Mukhopadhyay, R., Namboodiri, V.P., Jawahar, C.: A lip sync expert is all you need for speech to lip generation in the wild. In: Proceedings of the 28th ACM international conference on multimedia. pp. 484--492 (2020)

\bibitem{sak2014long}
Sak, H., Senior, A., Beaufays, F.: Long short-term memory based recurrent neural network architectures for large vocabulary speech recognition. arXiv preprint arXiv:1402.1128  (2014)

\bibitem{sinha2022emotion}
Sinha, S., Biswas, S., Yadav, R., Bhowmick, B.: Emotion-controllable generalized talking face generation. arXiv preprint arXiv:2205.01155  (2022)

\bibitem{song2022everybody}
Song, L., Wu, W., Qian, C., He, R., Loy, C.C.: Everybody’s talkin’: Let me talk as you want. IEEE Transactions on Information Forensics and Security  \textbf{17},  585--598 (2022)

\bibitem{suwajanakorn2017synthesizing}
Suwajanakorn, S., Seitz, S.M., Kemelmacher-Shlizerman, I.: Synthesizing obama: learning lip sync from audio. ACM Transactions on Graphics (ToG)  \textbf{36}(4),  1--13 (2017)

\bibitem{vaswani2017attention}
Vaswani, A., Shazeer, N., Parmar, N., Uszkoreit, J., Jones, L., Gomez, A.N., Kaiser, {\L}., Polosukhin, I.: Attention is all you need. Advances in neural information processing systems  \textbf{30} (2017)

\bibitem{vougioukas2020realistic}
Vougioukas, K., Petridis, S., Pantic, M.: Realistic speech-driven facial animation with gans. International Journal of Computer Vision  \textbf{128}(5),  1398--1413 (2020)

\bibitem{wang2020mead}
Wang, K., Wu, Q., Song, L., Yang, Z., Wu, W., Qian, C., He, R., Qiao, Y., Loy, C.C.: Mead: A large-scale audio-visual dataset for emotional talking-face generation. In: European Conference on Computer Vision. pp. 700--717. Springer (2020)

\bibitem{xie2007coupled}
Xie, L., Liu, Z.Q.: A coupled hmm approach to video-realistic speech animation. Pattern Recognition  \textbf{40}(8),  2325--2340 (2007)

\bibitem{zhou2019talking}
Zhou, H., Liu, Y., Liu, Z., Luo, P., Wang, X.: Talking face generation by adversarially disentangled audio-visual representation. In: Proceedings of the AAAI conference on artificial intelligence. vol.~33, pp. 9299--9306 (2019)

\bibitem{zhou2020makelttalk}
Zhou, Y., Han, X., Shechtman, E., Echevarria, J., Kalogerakis, E., Li, D.: Makelttalk: speaker-aware talking-head animation. ACM Transactions On Graphics (TOG)  \textbf{39}(6),  1--15 (2020)

\end{thebibliography}
\end{document}